# Scalable Formal Concept Analysis algorithm for large datasets using Spark

Raghavendra K Chunduri

Aswani Kumar Cherukuri
School of Information Technology & Engineering,
Vellore Institute of Technology, Vellore –India
E-mail: raghavkumar1988@gmail.com
E-mail: cherukuri@acm.org

**Abstract** In the process of knowledge discovery and representation in large datasets using formal concept analysis, complexity plays a major role in identifying all the formal concepts and constructing the concept lattice(digraph of the concepts). For identifying the formal concepts and constructing the digraph from the identified concepts in very large datasets, various distributed algorithms are available in the literature. However, the existing distributed algorithms are not very well suitable for concept generation because it is an iterative process. The existing algorithms are implemented using distributed frameworks like MapReduce and Open MP, these frameworks are not appropriate for iterative applications. Hence, in this paper we proposed efficient distributed algorithms for both formal concept generation and concept lattice digraph construction in large formal contexts using Apache Spark. Various performance metrics are considered for the evaluation of the proposed work, the results of the evaluation proves that the proposed algorithms are efficient for concept generation and lattice graph construction in comparison with the existing algorithms.

**1 Introduction**

Formal Concept Analysis was invented by Wille in early 80's. It is mainly used for analysis of object-attribute relationship and knowledge representation based on two notions: formal context(input)and formal concepts(output). The formal context is the input to FCA, which consists of set of objects, set of attributes and a binary relation that specifies which objects has which at- tributes[1]. Objects, attributes and the relation between objects and attributes is represented using a cross table, with rows representing the objects, columns representing the attributes and cross describing the relation between objects and attributes. Formal concepts are derived from the input formal context. Derived concepts are sorted in the order of inclusion and organized hierarchically to form a complete lattice called concept lattice[2]. Concept lattice is the core structure of FCA that formulates the knowledge and users can easily find the incidence relationship among objects and attributes[3]. The concepts in the concept lattice constitute a partial order relation that reflects the generalization and specialization within the concepts. There are various algorithms available in the literature for identifying concepts and constructing the concept lattice. The existing algorithms are divided into the batch and incremental algorithms. The batch algorithms identify the concepts quickly than the incremental algorithms because of a canonicity test that helps to avoid listing the same concept again[4]. For the same reason, in batch algorithms it is hard to build the incidence relationship among the concepts during the concept generation. Hence the concept lattice structure is not available explicitly in batch algorithms. Incremental algorithms maintain the incidence relationship among concepts hence they can obtain the concept lattice structure after concept generation. But the incremental algorithms does not have a canonicity test and it results in listing the same concept more than once, this increases the concept generation time. The problem of determining all concepts in the given formal context is #P-Complete[5]. The existing algorithms perform better for a smaller context and they are computationally weak, when they need to find the concepts in a large formal context and construct the lattice graph from the listed concepts.

In the recent Big Data era FCA is widely used as efficient data analysis technique. Several applications are using FCA for knowledge discovery and representation. For example in supervised learning some of the classification methods are proposed based on FCA [6]. Also in natural language processing FCA is used for learning concept hierarchies from text corpa [7]. With the extensive use of FCA in diverse fields, the complexity issues became a bottleneck for the effective use of FCA in all the domains. So an efficient FCA algorithm is needed for knowledge discovery (concept generation) and knowledge representation (concept lattice digraph construction) when dealing with large datasets. In this paper we propose an efficient distributed algorithm for concept generation and concept lattice construction using Apache Spark, a distributed in-memory processing framework used for iterative and interactive data analysis in large datasets. The proposed concept generation algorithm is a distributed batch processing algorithm that works iteratively for identifying concepts in the given large formal context. Then the lattice construction algorithm implemented using Spark Graphx module treats every concept as a node and constructs the digraph. The experimental analysis of the proposed work proves that the algorithm for concept generation is performing better than other existing distributed approaches while finding concepts. The algorithm completely utilizes the features of Apache Spark; the new canonicity test introduced in the proposed work is based on Spark storage level features. This test improved the performance of the algorithm significantly by eliminating the processing of duplicate concepts. Maintaining the parent index value for every concept helps in finding the incidence relationship among the generated concepts for the construction of the lattice graph.

The remainder of the paper is organized as follows. Section 2 recalls the basics definitions of FCA; Section 3 briefly introduces Apache Spark and its advantages over other existing Big Data processing frameworks. Section 4 gives a detailed note on the related work. In Section 5 we discuss the proposed work in detail. We demonstrate the experiments using the proposed method and presented the analysis of the results in section 6. Finally we conclude the paper with conclusion and directions for future work.

## 2 Formal Concept Analysis

In this section, we recall the basic definitions of FCA from Ganter and Wille. Then, we summarize the applications of FCA in diverse fields. The basics of FCA according to [1],[3] given as

Definition 2.1 A formal context is a triplet K = (G, M, I) where G is a non- empty set of objects, M represents a non-empty set of attributes, and I is a relation between G and M a subset of G * M Cartesian product. For the pair of elements

$$g \in G \text{ and } m \in M \text{ if}(g, m) \in I \qquad (1)$$

then this relation is expressed as object g has attribute m and writes as gIm. The derivation operators for a set A ⊆ G and B ⊆ M defined as

$$A^\uparrow = \{m \in M \text{ } | for\text{ } each\text{ } g \in A : (x, y) \in I\}, \qquad (2)$$

$A^\uparrow$ is the set of attributes common to the objects

$$B^\downarrow = \{g \in G \text{ } | for\text{ } each\text{ } m \in B : (x, y) \in I\}, \qquad (3)$$

$B^\downarrow$ is the set of objects common to the attributes

Definition 2.2 A formal concept of a context K (G, M, I) is a pair (A, B) defined as A ⊆ G and B ⊆ M, such that

$$A^\uparrow = B \text{ and } B^\downarrow = A \qquad (4)$$

A is called the extent and B is called the intent of the concept (A, B)

Defintion 2.3 The collection of all formal concepts in a given context K (G, M, I) order by ≤ is called concept lattice. A particular node in a lattice can be reached in various paths while hierarchical structures restrict each node to possess a single parent. The meet-mutual-sub-concept relation and the join- mutual-super-concept relation in a concept lattice is transitive that facilitates, a sub-concept of any given can be reached by traveling upwards from it. The top node in a concept lattice describes the generalization capability with all the objects of context in its extent. Similarly, the bottom node represents the specialization by exhibiting all the attributes of the context in its intent. From both generalization and specialization if an attribute-m(object-g) attached to a node in the lattice, then all the nodes below(above) must also contain the attribute-m(object-g). The lattice structure also gives both probabilistic and deterministic rules. The probabilistic rules are called as classification rules whereas the deterministic rules are known as functional dependencies.

Definition 2.4 An attribute implication represents an expression P → Q, where P, Q ⊆ M, is true in context K, if each object which has all attributes from P also has also all attributes from Q.

The number of possible implications that can be derived from a context can be exponential. For e.g. if the there are m number of elements in the attribute set M then $2^{2M}$ implications are possible from the context.

Defintion 2.5 A set X of attribute implications are called complete and sound with respect to a formal context K, if X is true in K and each implication is true in K follows from X.

Defintion 2.6 A set X on non-redundant attribute implications which is complete and sound with respect to a formal context K is a base with respect to a context K.

Basics of FCA models, applications and techniques are presented in the literature very nicely. FCA is widely used in diversified fields. Numerous applications are using FCA efficiently for effective results. Uta Priss has widely used FCA in natural language processing

applications[8][9]. Ignatov has given a detailed overview of FCA and its applications in information retrieval[10]. Kuznetsova and Poelmans have given a comprehensive literature review of FCA in vari- ous domains[11]. Aswani Kumar has proposed various methods for knowledge reduction in FCA[12][13]. FCA is also widely used in the field of Machine Learning. Sergei Kuznetsov has given a detailed overview on how FCA can be used in machine learning[14]. Ferrandin, Nievola, Enembreck, Ávila proposed a method for Hierarchical classification using FCA[15]. Patrick and Derek has proposed a collaborative filtering method using formal concept analysis[16]. Aswani Kumar and Premsingh has given a detailed overview on applications of FCA and its research trends[17]. With the extensive use of FCA in diverse fields, efficient FCA algorithms are required. Hence, we proposed algorithms for concept generation and concept graph construction using Apache Spark and tested against very large datasets.

## 3 Apache Spark

Distributed frameworks like MapReduce and its variants are popular and highly successful in implementing very large-scale data intensive applications on commodity hardware. However, these frameworks are not very suitable for iterative applications and application that requires to handle real-time data streaming. Hence, a new framework called Apache Spark has been proposed to overcome the drawbacks of MapReduce, using its underlying architecture. Spark is well suited for iterative applications that require access to the same data multiple times. The in-memory computations in apache spark increases the execution time ten times faster than the Hadoop MapReduce.

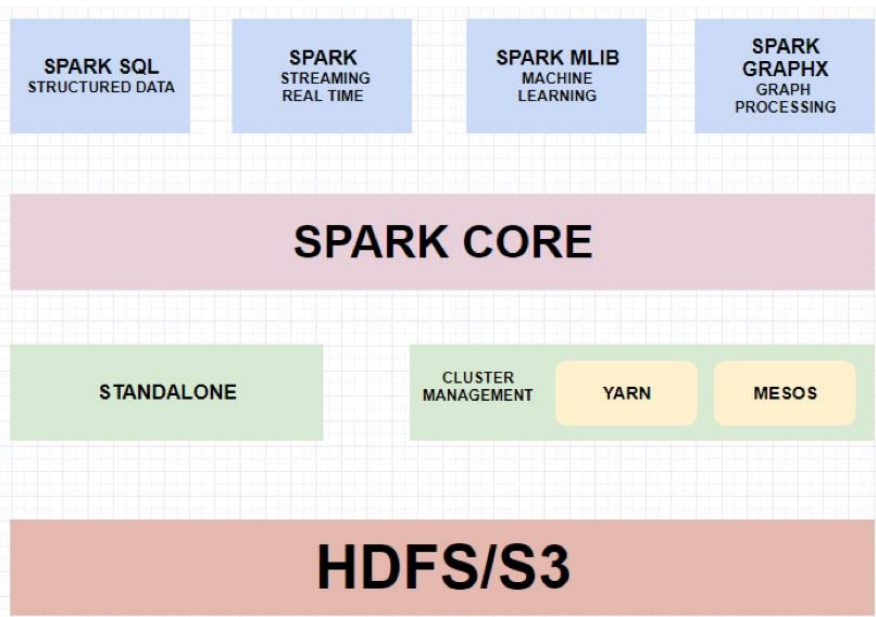

**Fig. 1** Spark Eco-System

Apache Spark is an open-source cluster computing framework that sup- ports flexible in-memory data processing that enables batch and real-time data processing. The idea behind the implementation of Spark is to develop a computing framework for distributed machine learning algorithms. Spark provides an interface for programming, entire clusters with inherent data-parallelism and fault tolerance[18]. Spark is integrated closely with other Big Data processing frameworks like Hadoop and accesses any Hadoop data sources while running. Spark abstracts the tasks of job submission, resource scheduling, tracking and communication between nodes, execution, and the low-level operations that are inherent in parallel data processing. Spark is used for a wide range of large-scale data processing tasks in machine learning and iterative analytics. Spark Core, Spark SQL, Spark MLlib and Spark Graphx, Spark streaming real-time are the main components of Apache Spark. Fig 1 rep- resents the Spark eco system. All the components of Spark eco system are discussed in the sequel.

3.1 Spark Core

Spark Core contains the basic functionality of Spark. Spark Core includes the components for memory management, task scheduling, interacting with storage systems and fault recovery. Spark Core also incorporates other features like resilient distributed dataset (RDD) a fault-tolerance collection of elements that operate in parallel. The data level parallelism in Spark is achieved through RDD's. More on Spark RDD are discussed in section 3.1.9

3.2 Spark SQL

Spark SQL is a package for working with structured data. It allows data querying via SQL and Hive, a variant of SQL called Hive Query Language (HQL). Spark SQL provides Spark with more information about the structure of both data and computations being performed. This information is mainly used to show extra optimizations. Spark SQL can be interacted using many ways, one of them is using SQL and Dataset API.

3.3 Spark Streaming

Spark Streaming is an extension of the Spark core API that enables scalable, high throughput and fault-tolerance processing of live data streams. Data for the Spark streaming can be ingested from various sources like S3, HDFS, Kafka, etc and the processed data can be pushed into file systems,
databases.

3.4 Spark MLlib

MLlib is Spark's machine learning library. The primary goal of MLlib is to make practical machine learning simple and scalable. MLlib implements common machine learning algorithms for classification, clustering, regression and collaborative filtering. According to the MLlib benchmarks, it is proved the MLlib is nine times faster than Apache Mahout.

### 3.5 Spark Graphx

Spark Graphx is a distributed graph processing framework used analysis of large graphs. Graphx performs parallel computations on graphs. Graphx API supports growing collection of graph algorithms and builders to simplify graph analytics and creation tasks. For constructing the digraph after generation of all the formal concepts we used Graphx. In the proposed work the concepts are attached to each vertex[19].

### 3.6 Driver Program

The driver program is the heart of Spark's Job execution process. The driver runs the application code that creates RDD's. The driver program creates SparkContext called the driver. The driver program splits the Spark applications into tasks and schedules, the tasks to run on executors in various worker nodes.

### 3.7 Spark Context

Spark context is another important component of Spark application and it is used as a client for Spark's execution environment. Spark context is used to get the current status of the application, it creates and manipulates the distributed datasets and manages the Spark's execution environment including running of Spark jobs, accessing services like task scheduler, block manager, etc.. The command used to create SparkContext is

val sc =new SparkContext(masternode, applicationName)

### 3.8 Spark Cluster Manager

Spark cluster manager plays a vital role in Spark's execution environment. The Spark context interacts with the cluster manager and gets the resources. It primarily acts like a Name Node in Hadoop. Spark cluster manager keeps track of all the resources like the number of worker nodes, the driver program, allocating resources, etc.. Fig.2. represents the detailed architecture of Apache Spark.

### 3.9 Spark Worker Node

Every worker node has executors running inside of them. Each executor will run multiple tasks. The tasks are the fundamental units of the execution. Every worker node also maintains the cache memory which plays a key role to improve the speed of apache Spark. Each worker node caches the data, and the cached data will be used for in-memory computations.

## 3.10 Spark Resilient Distributed Datasets(RDD)

RDD's are the Spark's core components. RDD's are the collections of data that are distributed and partitioned across all the nodes in a cluster. RDD's in Spark are fault-tolerant, this means that if a task on a given node fails, the RDD can be reconstructed automatically on the remaining nodes to complete the job. RDD's in Spark operates in parallel on its data[20]. RDD's in Spark can be created using the below snippets.

```
val inputRDD =sc.textFile(inputFile)
val inputRDD =sc.parallelize(list)
```

RDD's in Spark supports two types of operations, transformations and ac- tions. Examples of Transformations and actions are give below

Transformations : Map, FlatMap, Reduce etc..
Actions :count,reduce,saveAsTextFile etc...

Once we create the RDD in Spark, we can perform operations on the distributed dataset. These operations are split into transformations and actions. A transformation operation in Spark creates a new dataset from an existing one, where as an action operation returns a value by performing some computations on the dataset. An action operation returns its results to the driver program. Transformations in Spark are lazy, they do not compute their results right away. The transformations are computed when an action requires a result that needs to be sent to driver program. Each RDD maintains a pointer to one or more parents along with the metadata about, that is the type of relationship it has with its parent. When we call a transformation on RDD, the RDD just keeps a reference to its parent, and it is called lineage. Spark creates a lineage graph with all the series of transformations that are applied. When the driver submits the job, the lineage graph is serialized to the worker nodes. Each of these worker nodes applies transformations on different nodes and allows an effective fault-tolerance by logging all the transformations that are used to build the dataset. If a partition of RDD is lost, the RDD has enough information from the lineage graph to recompute the failed partition, thus lost data can be recovered quickly, without requiring the costly computation again.

## 3.11 Caching RDD's

The most important feature of the Spark is its ability to cache the data in memory across the cluster. The cache method is used to cache the data and tells Spark that RDD should be kept in memory. The first time when an action is called on RDD that initiates a computation, the data is read from the disk and stores into memory. Hence for the first time such an operation is initiated, the time it takes to run the task is more because of reading the input from disk. From the second time, when the same data is accessed the data can be read directly from cache memory which avoids expensive I/O operation and speeds up the processing and improves the performance of the algorithm significantly.

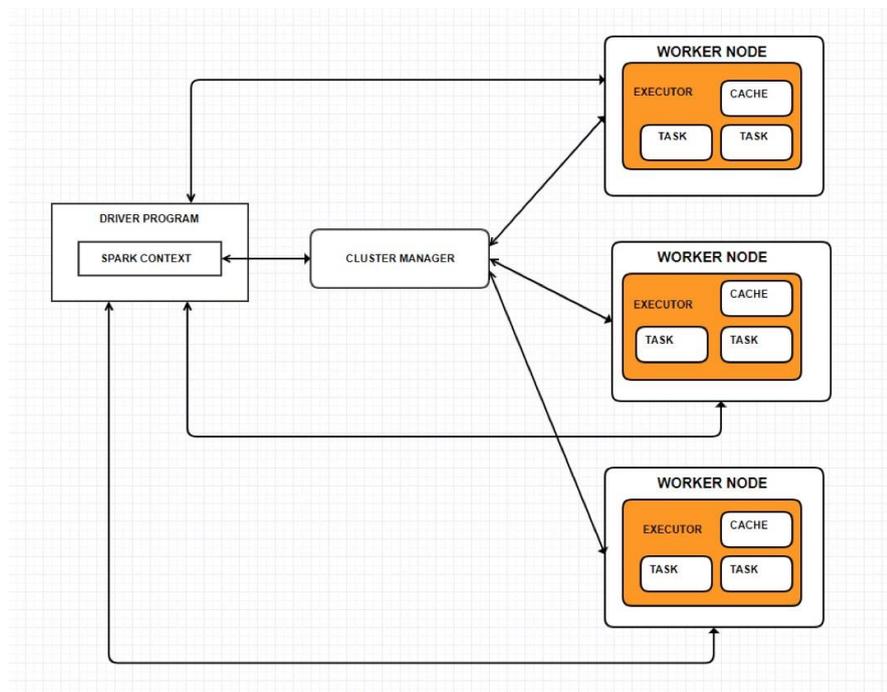

**Fig. 2** Architecture of Apache Spark

## 4 Related Work

Several Algorithms are proposed for concept generation and concept lattice construction [21],[22],[23],[24],[25]. All these algorithms are best suitable for execution on a single node cluster and are not efficient enough to handle large datasets. There are few concept generation algorithms implemented using MapReduce and its iterative variants like HaLoop and Twister. However these algorithms are feasible for larger datasets, but they are not efficient enough while dealing with large datasets because of the bottlenecks in MapReduce. All the existing algorithms that adopted MapReduce framework have not addressed the bottlenecks of MapReduce like efficient utilization of CPU and memory, in-memory computations and disk I/O after each phase in MapReduce.

Biao Xu, Ruairı de Frein, Eric Robson, and Mıcheal O Foghl[26] proposed an iterative concept generation algorithm using Twister; a lightweight iterative runtime environment for iterative MapReduce applications. The algorithms are the implementations of Ganter next concept algorithm using iterative MapReduce framework and the algorithms are called MRGanter and MRGanter+. However the iterative approach in twister is not well designed to handle fault tolerance and failures effectively. A single failure will result in executing the current iteration again from the beginning, irrespective of the level of completion of the iteration before the failure occurs. This increases the execution time in case of failures. Also in twister, it is hard to eliminate the shuffling, sorting and grouping of redundant data. In the Ganter's algorithm the concept lattice structure is not available immediately because the lattice is an implicit property of the generated concepts.

Nilander, Sergio, Henrique and Luis[27] proposed a parallel algorithm implemented using OpenMP based on Ganter's next closure algorithm. OpenMP is an api that uses multithreading and executes the algorithm using threads and shared memory. OpenMP is not feasible to larger datasets because of its architecture complexity. In shared memory systems all the data needs to be loaded into the shared memory and this is really a problem when working on datasets with high dimensionality. Authors also did not discuss on the processing of duplicate concepts, how the CPU and memory are utilized efficiently while running threads on the CPU cores. There are no fault tolerance techniques in OpenMP architectures, which is really hard to come to a stable state after the recovery from a failure.

[28]Bhatnagar and Lalit proposed a MapReduce implementation of concept generation. This algorithm performs computations at reduce phase. This algorithm is not able to find all the formal concepts for the given formal concept. Only a sufficient set of concepts are identified during the single iteration. Authors did not mention how the generated concepts are sufficient and how can the existing concepts generate the remaining concepts when needed.

[29][30] RKChunduri, CA Kumar proposed an approach using HaLoop, an- other iterative MapReduce framework. HaLoop takes a large number of iterations to process the larger datasets. HaLoop has a cache memory concept but it does not support in-memory computations. Since HaLoop adopts most of the architecture model from Hadoop and uses MapReduce, the bottlenecks in the MapReduce system still exists in HaLoop based concept generation.

The below table shows the various properties of various distributed algorithms including the proposed work.

All the algorithms discussed in the above table are batch processing algorithms and they don't possess incidence relationship among the generated concepts. Without an incidence relationship among the generated concepts it is hard to construct the lattice after concept generation. All the MapReduce algorithms discussed in the above table except HaLoop implementation uses fault tolerance of Hadoop framework. In HaLoop there is an extra care taken to handle fault tolerance to avoid the iterations to start from scratch after a failure and the OpenMP architecture does not possess any fault tolerance. So, to overcome all the drawbacks in the existing work, the proposed algorithms efficiently uses the features of Spark to improve the concept generation process.

**Table 1** Properties Comparison of various distributed concept generation algorithms

| Algorithm | Incidence Relationship | Canonical test | in-memory computation | Cache Support | Fault tolerance |
|---|---|---|---|---|---|
| Twister Based MapReduce concept generation | no | no | no | no | yes |
| HaLoop Based MapReduce concept generation | no | no | no | yes | yes |
| Parallelization of NextClouser using OpenMp | no | no | no | no | no |
| MapReduce algorithm for computing formal concepts in Binary data | no | no | no | no | yes |
| SparkConcept Genera- tion algorithm | yes | yes | yes | yes | yes |

There are few concept lattice construction algorithms in the literature, but they are not distributed in nature. Muangprathub[31] proposed a novel algorithm for building concept lattice, which depends on the size of the extent for lattice construction. They purely depend on the size of the extent for calculating the concept level; there are chances that different sizes of the concept extent may sometime fit in the same level, which is not addressed in this approach. This algorithm takes high time to construct the lattice with a large number of formal concepts because of its sequential approach.

**5 Proposed Work**

In this section, we discuss the proposed work by dividing it into two stages. The first stage identifies the formal concepts for the input formal context, the proposed algorithm is named as SparkConceptGeneration algorithm. The digraph construction of the takes place in the second stage and the algorithm is named as SparkLatticeConstruction algorithm The pseudo code of the two algorithms is given below with a detailed explanation.

5.1 **Explanation of SparkConceptGeneration Algorithm**

The SparkConceptGeneration algorithm is a distributed algorithm formalized by a recursive function NeighborConcept (), which lists all the formal concepts starting from the least formal concept. The recursive function NeighborConcept() takes tuple called concept as an input that has 4 parameters, the extent and intent of the concept, isValidNeighbor a boolean flag, and parent index. The parameter isValidNeighbor is used to determine whether a particular input concept can generate the neighbor concepts or not. This parameter is useful for the first step in the canonicity test. The parameter parent index determines the level of the concept in lattice graph. The value of the parent index is set to '1' for the least formal concept and the value gets incremented for the neighbor concepts that are generated from every concept. The NeighborConcept () function generates all the upper neighbor concepts in different iterations and stops after the greatest formal concept is found. Every iteration of the recursive process undergoes a two step canonicity test to make sure that the generated concepts are not considered for processing again. The concepts generated in each iteration will be saved into Spark's RDD and persisted into cache. The second step of the canonicity test will check the cache before processing the concept and decides whether to take up the concept for processing or not. The SparkConceptGeneration algorithm along with the two-step canonicity test is presented below. The SparkConceptGeneration algorithm starts by creating a Spark context and takes formal context file as an input. The objects and attributes in the input file will be separated by a character which is a run time parameter decides based on the type of data. In this paper we have taken"," as a separator and formulated the pseudo code using",". The below pseudo code returns a tuple that has objects and attributes of the context. They are saved into RDD's called contextObjectsRDD and contextAttributesRDD using RDD actions.

```
1 val sc = new SparkContext(new SparkConf().setAppName("appName"));
2 val context = sc.textFile(inputFile).map {
 line => val data = line.split(",");
3 val contextObjectsRDD = data.head ;
4 val attributesRDD = data.tail; }
```

The flowchart in fig 3 represents the execution flow of neighbor concept recursive function presented in algorithm 1.

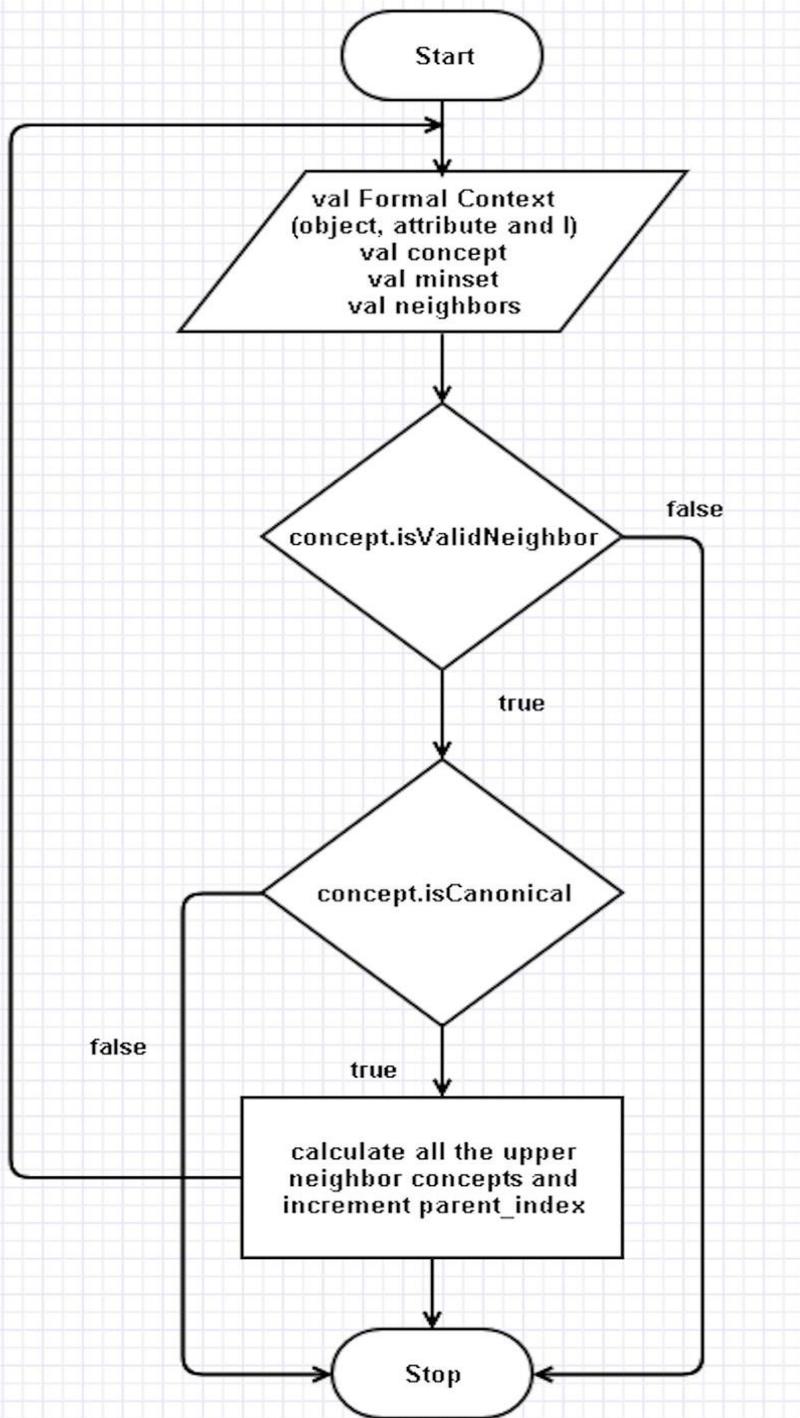

**Fig. 3** Execution flow of recursive function NeighborConcept

The overview of the SparkConceptGeneration algorithm presented in algorithm 1, is discussed below.

Step 1: Initially the least formal concept is calculated, and function NeighborConcept () will be called with the least formal concept. For the least formal concept the isValidNeighbor flag is set to true and the parent index is set to 1.

Step 2: The recursive function NeighborConcept () is a recursive function lists all the neighbor formal concepts for the least formal concepts and increments the parent index by 1 for the generated neighbor concepts.

Step 3: Two step canonicity tests will be performed for each of the concept generated and finds the upper neighbor concepts for every concept listed.

Step 4: The isValidNeighbor flag for each concept is checked during the first step of the canonicity test. If the value of this flag is true, then the cache will be checked to see whether the concept is present in it or not. If it is presented in the cache then the next concept in the RDD will be picked.

Step 5: The steps 1 through 4 will be repeated until the greatest formal concept is reached.

The Pseudo code for SparkConcept algorithm explained above, works as follows

---

**Algorithm 1:** SparkConceptGeneration

**Input :** *Concept concept*
**Output:** *neighbor*

```
1  NeighborConcept(Concept concept){
2  val minSet = Set(contextObjectsRDD) - Set(concept.extent)
3  val minRDD =contextObjectsRDD - concept.extent
4  val neighbors = ∅
5  if concept.IsValidNeighbor then
6      if isCanonical (concept.intent) then
7          parent_index =parent_index +1
8          for x in (contextObjectsRDD - (concept.extent)) do
9              B1 = concept.e (objectConceptFormingOperator(x))
10             A1 = attributeConceptFormingOperator(B1)
11             if minRDD∩((A1(concept.extent))-(x) == ∅) then
12                 neighbors =neighbors ∪ (Concept(A1, B1, true, B1.size, parent index))
13                 neighbors.cache
14             else
15                 minSet= minSet - x.toSet
16                 neighbors =neighbors ∪ (Concept(A1, B1, false, B1.size, parent index))
17                 neighbors.cache
18             end
19         end
20     end
21  end
22  for neighbor in neighbors do
23      NeighborConcept(Concept concept)
24  end
25  }
```

The steps 1, 2, 3 and 4 in algorithm 1 are the initialization steps, followed by a two step canonical test in steps 5 and 6. The step 5 checks for isValidNeighbor element of each concept to find whether the given concept is able to generate any neighbors or not. If the isValidNeighbor is not able to generate any concepts, then next concept in the RDD will be picked. If a concept is able to generate neighbors then the intent of the concept is checked to make sure that the algorithm is not processing already listed concept. The isCanonical () method in algorithm 4 checks the cached concepts and returns true if the intent of the concept not matches any of the concepts intent in the cache. If it matches it will return false. Even if the second step of the canonical test fails, the SparkConceptGeneration algorithm proceeds with the next concept. After the greatest formal concept is reached the algorithms stops listing the concepts and writes all the generated concepts into a file. The programming model of SparkConceptGeneration is shown in Fig 4 .While listing all the formal concepts, the context forming operators and are required. The pseudo code for calculating the context forming operators is discussed below in algorithm 3 and 4. For calculating the an d the context and its inverse has to be converted into map. The below pseudo code construct the map from the given formal context and its inverse.

```
1 val contextAsMap = sparkContext.map
2 line =>
3 val data = line.split(",")
4 val attributes = data.tail.mkString(",").split {(",").map{ f =>
5 val attribute = f
6 (attribute, data.head)
7  (attributes)}.collectAsMap
```

```
1 val contextInverseAsMap = contextAsMap.map{
2 f => val contextInverseListMerge = f
3 (contextInverseListMerge)
4}.reduce(_ union_).group By( _._1).map { case (k, v) => (k,
v.map( _._2).toSet) }
```

**Algorithm 2:** AttributeConceptFormingOperator
1  attributeConceptFormingOperatorconcept.object
2 *Input* : *attribute*
3 *Output* : *allobjectssharingtheattribute*
4 val objectRDD=sc.parallelize(contextAsMap.get(concept.attribute).toSeq)
5 return objectRDD

**Algorithm 3:** ObjectConceptFormingOperator
1 objectFormingOperatorconcept.attribute
2 *Input* : *object*
3 *Output* : *allattributessharingtheobject*
4 val attributeRDD = sc.parallelize(contextInverseAsMap.get(concept.attribute).toSeq) return attributeRDD

Second step in two step Canonical test

**Algorithm 4:** IsCanonical
1 isCanonical(concept.intent)
2 *Input* : *intentoftheconcept, neighborsRDD*
3 *Output* : *true/false*
4 val neighbors =getPersistentRDDs()
5 val count =neighbors.filter(. 2.contains(concept.intent)).count()
6 if *count > 0* then
7    false
8 else
9    true
10 end

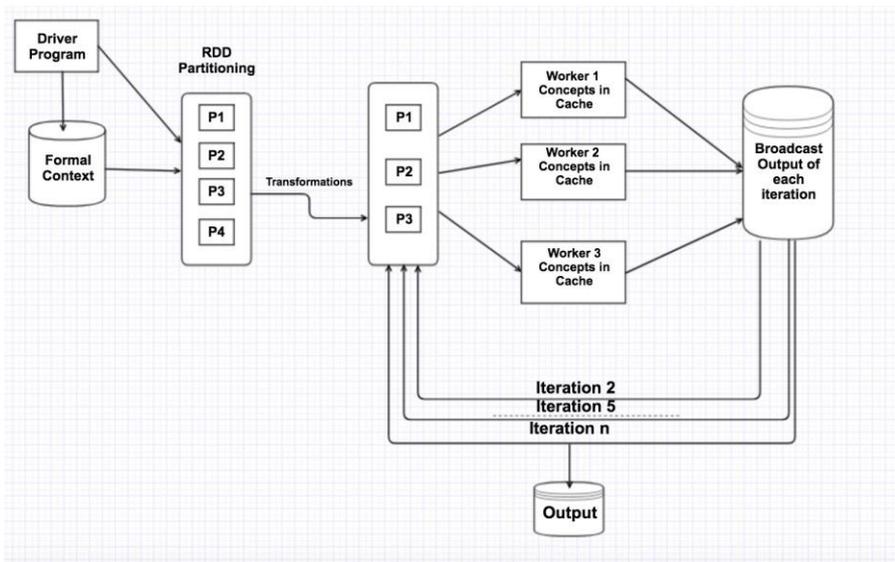

**Fig. 4** Programming model of Concept Generation using Spark

The SparkLatticeConstruction algorithm takes the concept file as an input and constructs the concept digraph using Spark Graphx module. The pseudo code for the SparkLattice method is discussed in algorithm 5.

1. val sc = new SparkContext(new SparkConf().setAppName("graph-generation"));
2. val conceptsRDD = sc.textFile(NeighborConceptsFile).map
3. line =>(Concept(line)) }

The above pseudo code takes the concepts file as an input and converts each line of the file to a concept tuple and stores it into an RDD. Now the concepts are sorted based on the parent index and zipped with a unique index using the zipwithIndex method. A vertex table is constructed by taking zipped index as vertex id and the concept tuple as its property. Now the difference between the parent index between each concept will be calculated, and if the difference is equal to 1, then an entry into the edge table will be added. This process identifies all the edges in the graph and constructs a digraph of concepts using the vertex table and edge table.

**Algorithm 5:** SparkLatticeConstruction Algorithm
1. *Input* : *NeighborConceptsFile*
2. *Output* : *Graph B*(*G, M, I*)
3. val vertextTable,edgeTable
4. vertexTable == conceptsRDD.sortBy(f =>f. 5, true).zipwithIndex
5. conceptsRDD.reduce{
6. f => (a,b) => val difference ==(a._5 - b._ 5))
7. if *(difference ==1)* then
8. edgeTable = edgeTable.add(a,b)
9. end
10. Graph B =(vertextTable,edgeTable)

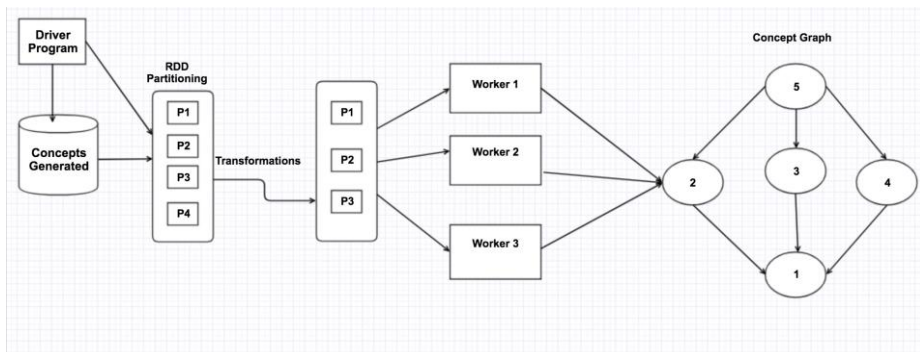

**Fig. 5** Programming model of SparkLatticeConstruction algorithm

Fig 5 explains the programming model of Spark lattice construction algorithm. The concepts generated by the Spark concept generation algorithm are taken from the file and a RDD of concepts are created. Now the concept RDD is distributed among workers to determine vertices and edges of the graph. Finally the digraph of the concepts is constructed which is the desired output.

The proposed algorithms for generating the formal concepts has the time complexities $G^2 * M$. The canonical test to verify the intent of the concepts in the cache is GM which can be ignored. The SparkLattice algorithm for building the graph has the time complexity $v*e^2$ where v is the total number of vertices and e is the total number of edges.

## 6 Experimental Analysis

We have implemented the SparkConceptGeneration and SparkLatticeConstruction algorithms using Apache Spark Scala API. The experiments were run on the Google cloud cluster with various configurations of worker nodes ranging from 1-16 and tested on the various datasets from UCI machine learning repository. The three datasets considered for the experiments ranging from small car evaluation dataset that has 1728 objects to large Poker-Hand dataset that has half a million objects. We carried out the experiments five times on the clusters using datasets with worker nodes ranging from 1 to 16 and considered the average of the experiments because of the reason that the single time execution experiment results produced different results every time we execute the algorithm. The best of the results were observed, when the algorithm's executed on 1, 4, 10, 16 node clusters. In formal concept analysis generating concepts from the given formal context is the important step, so most of the evaluations are based on SparkConceptGeneration algorithm and compared it with the other distributed works implemented using MapReduce. The following parameters are evaluated as part of the experimental analysis.

6.1 Execution time

Execution time is one of the performance metrics where we measured the time required for generation of concepts and construction of the concept lattice. The table represents the datasets, the number of objects, attributes in the dataset, the number of nodes and the execution time to process each dataset for generation of concepts. The results are depicted for each experiment and the average of the results is given in the last row of the table. For all the datasets, with an increase in the number of worker nodes, the execution time is significantly decreasing. A detailed comparison of the proposed work with other distributed implementations is discussed in the sequel.

Tables 2 to 5 represents the execution times of the SparkConcept algorithm on a cluster that has 1, 4, 10, 16 nodes. On each of the cluster instance the algorithm is executed on 3 input datasets and the results are observed.

**Table 2** Execution time in seconds for datasets on the cluster that has one node

| Experiment Number | Car Evaluation dataset Objects: 1728 Attributes:6 | Adult dataset Objects : 32561 Attributes:15 | Poker Hand dataset Objects: 512505 Attributes: 11 |
|---|---|---|---|
| 1 | 623 | 2435 | 7246 |
| 2 | 645 | 2525 | 7389 |
| 3 | 617 | 2489 | 7273 |
| 4 | 619 | 2396 | 7216 |
| 5 | 605 | 2461 | 7257 |
| Average | 621 | 2462 | 7276 |

**Table 3** Execution time in seconds for datasets on the cluster that has four nodes

| Experiment Number | Car Evaluation dataset Objects: 1728 Attributes:6 | Adult dataset Objects : 32561 Attributes:15 | Poker Hand dataset Objects: 512505 Attributes: 11 |
|---|---|---|---|
| 1 | 219 | 1054 | 3315 |
| 2 | 275 | 1021 | 3372 |
| 3 | 246 | 1103 | 3345 |
| 4 | 197 | 1067 | 3340 |
| 5 | 203 | 1057 | 3329 |
| Average | 228 | 1060 | 3340 |

**Table 4** Execution time in seconds for datasets on the cluster that has ten nodes

| Experiment Number | Car Evaluation dataset Objects: 1728 Attributes:6 | Adult dataset Objects : 32561 Attributes:15 | Poker Hand dataset Objects: 512505 Attributes: 11 |
|---|---|---|---|
| 1 | 86 | 374 | 1367 |
| 2 | 85 | 335 | 1392 |
| 3 | 93 | 363 | 1384 |
| 4 | 95 | 372 | 1349 |
| 5 | 81 | 398 | 1326 |
| Average | 88 | 369 | 1363 |

**Table 5** Execution time in seconds for datasets on the cluster that has 16 nodes

| Experiment Number | Car Evaluation dataset Objects: 1728 Attributes:6 | Adult dataset Objects : 32561 Attributes:15 | Poker Hand dataset Objects: 512505 Attributes: 11 |
| --- | --- | --- | --- |
| 1 | 36 | 123 | 656 |
| 2 | 32 | 136 | 643 |
| 3 | 40 | 146 | 653 |
| 4 | 32 | 128 | 682 |
| 5 | 35 | 133 | 650 |
| Average | 35 | 133 | 657 |

The below figure 6 shows the average values (presented in last row of Table's 2-5) of the experiments when the algorithm is executed on different datasets. From the above graph we can observe that the algorithm is generating concepts quickly with the increase in the number of nodes, i.e., with the increase in the number of executors that works in parallel increases, the execution time is decreasing. The execution time is also depending on the size of the dataset. For smaller datasets, the concepts are generated in a short span. For car evaluation dataset on 16 node cluster the algorithm has generated 35 concepts in 16 seconds, and for Poker hand dataset it took 657 seconds for generation of 148726 concepts which is better than the MapReduce approach.

The table 6 shows the number of concepts generated for each dataset and the time it took for the construction of the lattice graph. With the increase in the number of nodes, the time for constructing the graph is decreasing. The vertex table and edge table constructed in the SparkLatticeConstruction algorithm are RDD's that are distributed across the worker nodes for distributive construction of the graph.

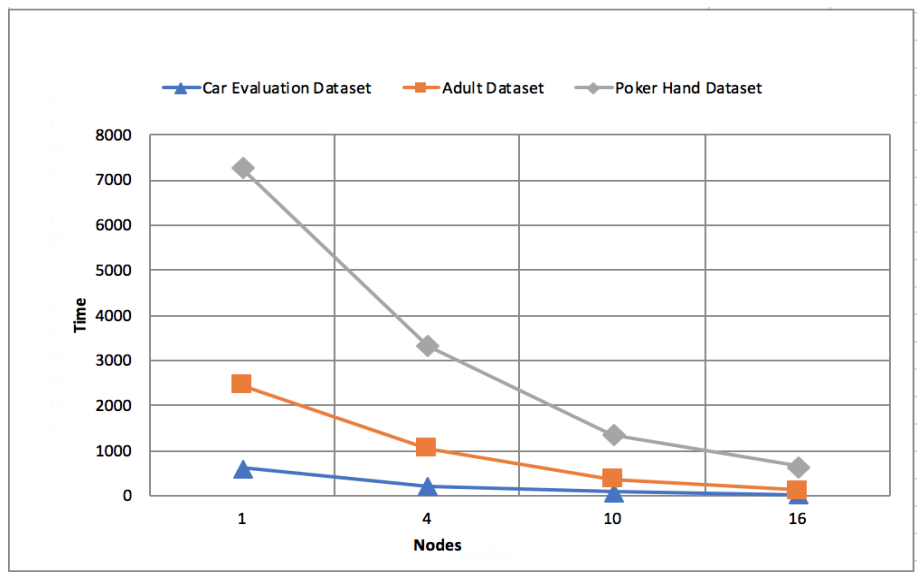

**Fig. 6** Average time in seconds for SparkConceptGeneration on all cluster that has , 10,16 worker nodes in a cluster

**Table 6** Number of Concepts generated and time taken lattice graph construction on each node

| Dataset | Number of Concepts Generated | 1 Node Cluster | 4 Node Cluster | 10 Node Cluster | 16 Node Cluster |
|---|---|---|---|---|---|
| Car Evaluation | 35 | 746 | 383 | 132 | 86 |
| Adult | 12678 | 2976 | 1672 | 763 | 345 |
| Poker Hand | 148726 | 5634 | 2996 | 1532 | 578 |

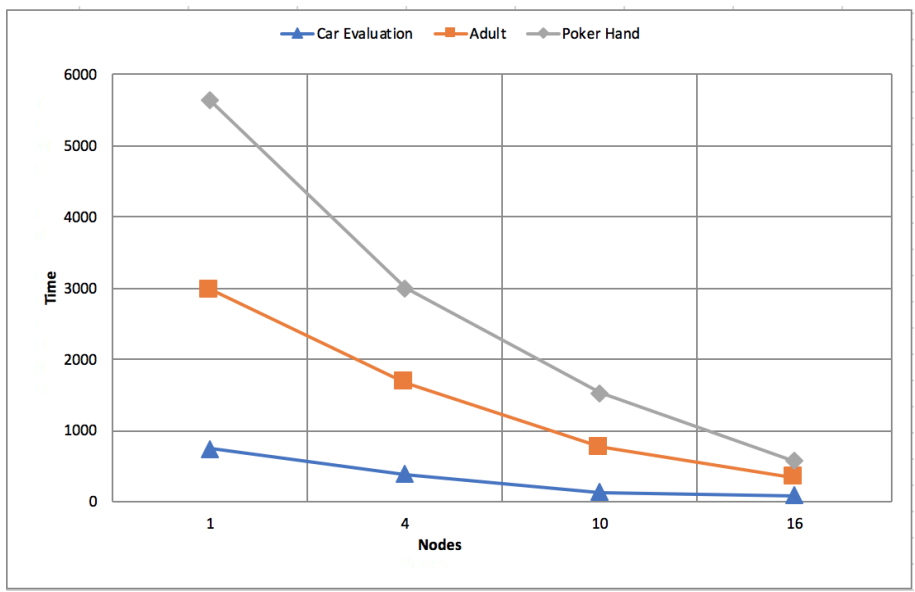

Time in seconds for SparkLattice construction on all Datasets on 1,4,10,16 node cluster

From the above graph in Fig 7 we understand that the time taken for construction of lattice graph from the car evaluation dataset concepts on a 16 node cluster is 86 seconds and for the Poker-Hand dataset it is 578 seconds.

## 6.2 CPU Utilization

CPU utilization is one of the important performance metrics; the results show the average of maximum CPU utilization for generating all the concepts in comparison with the MapReduce approach for concept generation.

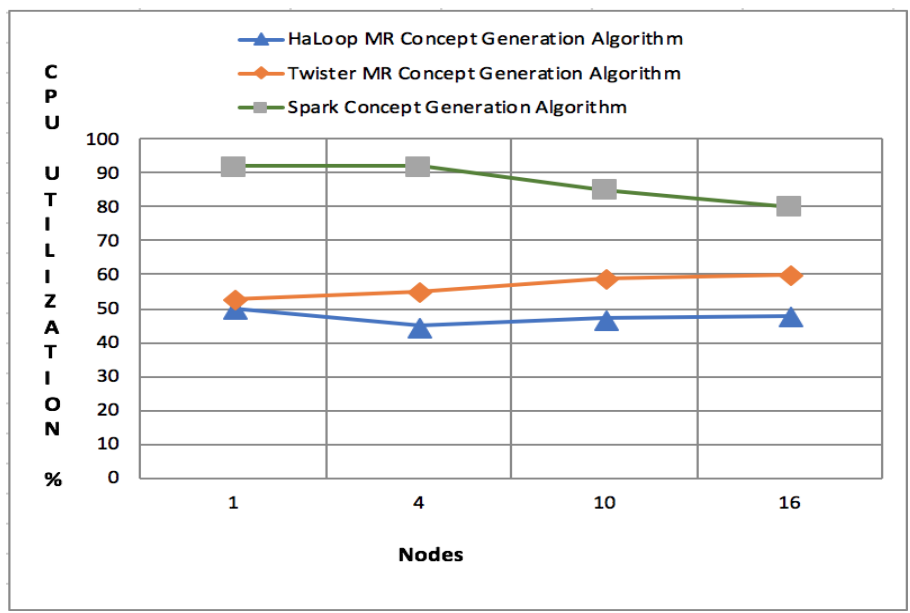

**Fig. 7** CPU utilization of Spark and MapReduce based concept generation algorithms while executing on 1, 4, 10, 16 node cluster

The above graph in Fig 8 shows that the Spark is efficiently using CPU's for its application processing. With the increase in the number of nodes, the number of CPU cores will be increased. For SparkConceptGeneration algorithm, on a single node cluster, the CPU is completely occupied with the application tasks. With the increasing nodes, more CPU is available for utilization. The two MapReduce algorithms have not utilized more than 60% of CPU's. In the MapReduce implementation, every TaskTracker has the map and reduce slots which are not generic slots. When a MapReduce applications start it may spend hours of time in map phase, during this time the reducer slots are not in use, and they are idle. Because of this reason the CPU% is not very high because of the empty reducer slots. Spark has the concept of tasks which are generic and always tries to utilize the maximum CPU, because of this reason CPU utilization is almost 80% in all cases in the SparkConceptGeneration algorithm and proved that the Spark is efficiently using the CPU's resources in the cluster to improve the performance of the algorithm.

Fig 9 represents the CPU utilization of the Spark Concept generation algorithm while executing on a Google cloud cluster that has 4 worker nodes. The maximum CPU% utilized on this cluster is 96.67%.

**Fig. 8** CPU utilization of SparkConceptGeneration algorithm on a 4 Node cluster

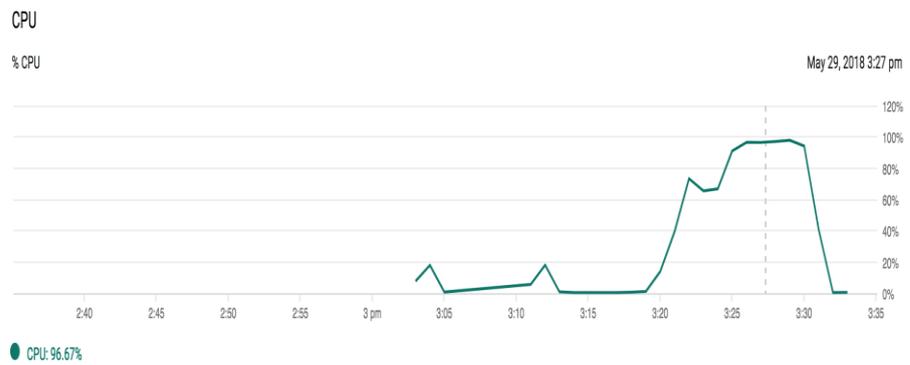

6.3 Memory Utilization

Memory Utilization is one of the vital performances metrics that needs to be considered. The results in Fig 10 shows that the Spark is utilizing maximum memory in every case. The cluster is configured with 64GB of memory on each node and Spark utilized 90% of the memory. Spark is an in-memory processing engine that keeps all of the data, and intermediate results in memory. In-Memory storage is the main reason for the faster speed of spark while processing. In the Map Reduce, the intermediate results are written to the HDFS (disk every time) and read back the data again from HDFS. The memory utilization in MapReduce algorithms is less because of the lack of in-memory computations. The HaLoop based Map reduce approach stores data in the cache, but it will not do any computations. HaLoop uses memory more than the twister environment because of the various cache it is supporting. Fig 11 represents the memory utilization of the SparkConceptGeneration algorithm while executing on a google cloud cluster that has 4 worker nodes and 30GB of memory. The maximum memory utilized for in-memory processing is around 22GB and the cache memory utilized for persisting the generated concepts is around 350MB.

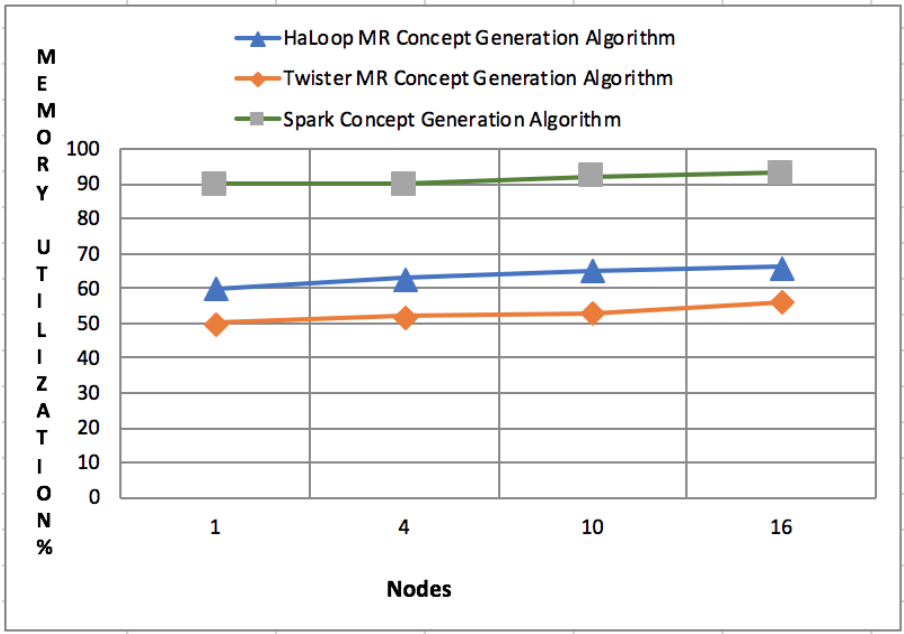

**Fig. 9** Memory utilization of Spark and MapReduce based concept generation algorithms while executing on 1, 4, 10, 16 node cluster

```
raghavkumar1988@cluster-7059-m:~$ free -m
             total       used       free     shared    buffers     cached
Mem:         30225      22304       7920          8         20        350
-/+ buffers/cache:      21933       8291
Swap:            0          0          0
```

**Fig. 10** Memory utilization of SparkConceptGeneration algorithm on a 4 Node cluster that has 30GB memory

   The execution times of our proposed algorithms are compared with the existing related works. There are no works in the literature that used Apache Spark for concept generation, so we have considered the MapReduce implementations of concept generation for the analysis and observed that in all the cases the Spark implementation is performing better than the existing approaches. The underlying architecture of Spark significantly reduced the number of iterations to generate concepts when compared with iterative MapReduce approaches like HaLoop and Twister. The in-memory computations greatly reduced the concept generation time by generating large number of concepts within a short span. Tables 7 to 10 represents the execution times of all the three approaches (Spark, Twister and HaLoop) when executed on different ranges of cluster using all the three input datasets.

**Table 7** Execution times of Spark, HaLoop and Twister approaches in seconds on the cluster that has one nodes

| Algorithm | Car Evaluation dataset Objects: 1728 Attributes:6 | Adult dataset Objects: 32561 Attributes:15 | Poker Hand dataset Objects: 512505 Attributes: 11 |
|---|---|---|---|
| HaLoop MR Concept Generation Algorithm | 1945 | 5782 | 13455 |
| Twister MR Concept Generation Algorithm | 2347 | 6683 | 18796 |
| Spark Concept Generation Algorithm | 621 | 2462 | 7276 |

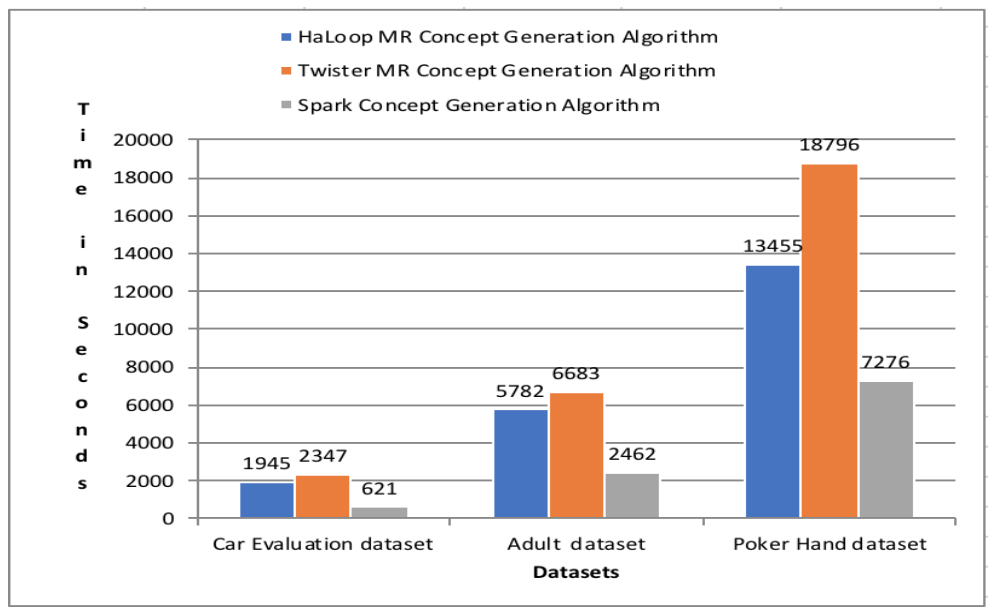

**Fig. 11** Comparison of Spark and MapReduce implementations on single node cluster

**Table 8** Execution times of Spark, HaLoop and Twister approaches in seconds on the cluster that has four nodes

| Algorithm | Car Evaluation dataset Objects: 1728 Attributes:6 | Adult dataset Objects : 32561 Attributes:15 | Poker Hand dataset Objects: 512505 Attributes: 11 |
|---|---|---|---|
| HaLoop MR Concept | 1056 | 2913 | 10765 |
| Twister MR Concept Generation Algorithm | 1176 | 4524 | 12315 |
| Spark Concept Generation Algorithm | 228 | 1060 | 3340 |

**Table 9** Execution times of Spark, HaLoop and Twister approaches in seconds on the cluster that has ten nodes

| Algorithm | Car Evaluation dataset Objects: 1728 Attributes:6 | Adult dataset Objects : 32561 Attributes:15 | Poker Hand dataset Objects: 512505 Attributes: 11 |
|---|---|---|---|
| HaLoop MR Concept Generation Algorithm | 835 | 1926 | 7821 |
| Twister MR Concept Generation Algorithm | 1263 | 2198 | 8326 |
| Spark Concept Generation Algorithm | 88 | 369 | 1363 |

**Table 10** Execution times of Spark, HaLoop and Twister approaches in seconds for datasets on the cluster that has 16 nodes

| Algorithm | Car Evaluation dataset Objects: 1728 Attributes:6 | Adult dataset Objects : 32561 Attributes:15 | Poker Hand dataset Objects: 512505 Attributes: 11 |
|---|---|---|---|
| HaLoop MR Concept Generation Algorithm | 679 | 1375 | 5074 |
| Twister MR Concept Generation Algorithm | 854 | 2334 | 5983 |
| Spark Concept Generation Algorithm | 35 | 133 | 657 |

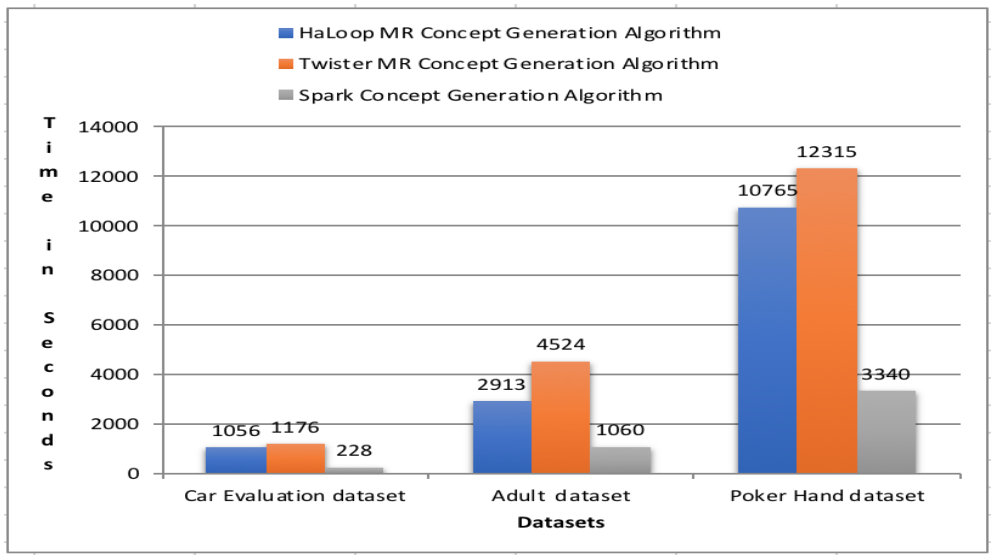

**Fig. 12** Comparison of Spark and MapReduce approaches on four node cluster

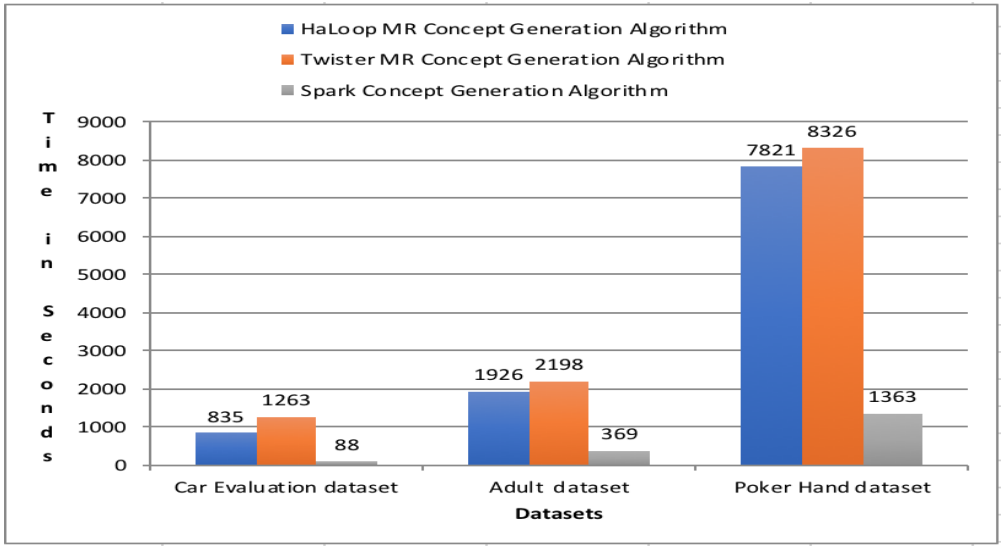

**Fig. 13 Comparison** of Spark and MapReduce approaches on 10 node cluster

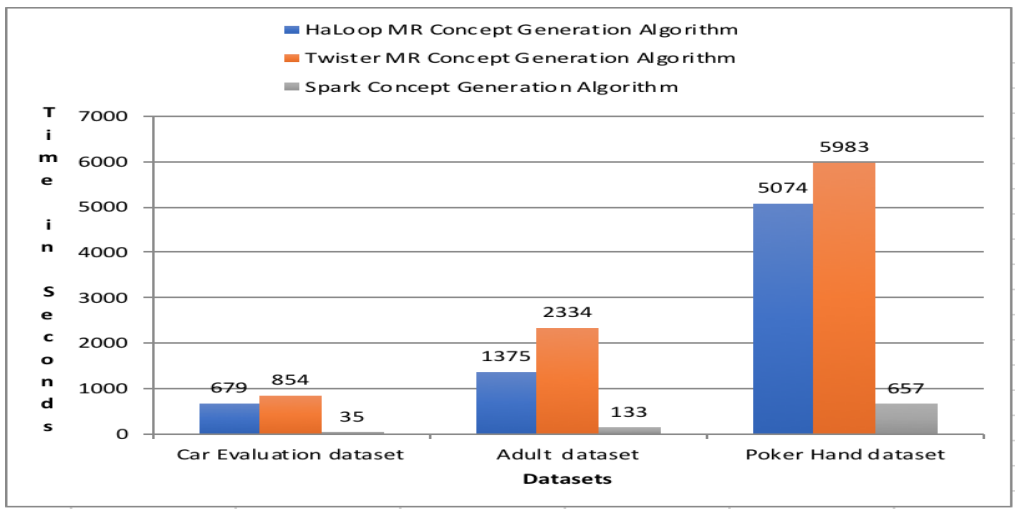

**Fig. 14** Comparison of Spark and MapReduce approaches on 16 Node cluster

All the algorithms for concept generation considered for evaluation are implemented using HaLoop and Twister based MapReduce environments. These algorithms are executed on 1 Node, 4 Node, 10 Node and 16 node clusters and the graphs in Fig 12 to 15 represents the comparison of the three approaches . In all the case Spark implementation performed better than the other two approaches. The main reasons for the better performance of Spark is due to its data-level parallelism and in-memory computations. Spark runs its job by spawning different threads running inside the executor. A thread is a lightweight process that runs part of the task, whereas in MapReduce the map and reduce processes are heavyweight. Spark extensively utilizes the CPU and memory, where as MapReduce implementations failed to do so because of the architectural complexity. The below graph in Fig 16(values are presented in Table 11) shows the number of iterations that all three algorithms underwent when executed on 16 node cluster. The numbers of iterations in Spark are decided based on RDD partitions and size of the memory. Spark takes less number of iterations when compared to the number of iterations in HaLoop and Twister because of data-level parallelism in RDD's. In HaLoop and Twister environments the number of iterations will be decided based on the size of data block which is 64 MB. The main reason for more number of iterations in MapReduce is because of the complex architecture, and each iteration runs as an independent MapReduce job. HaLoop and Twister are still using the underlying architecture of Hadoop which is not a good design for iterative applications.

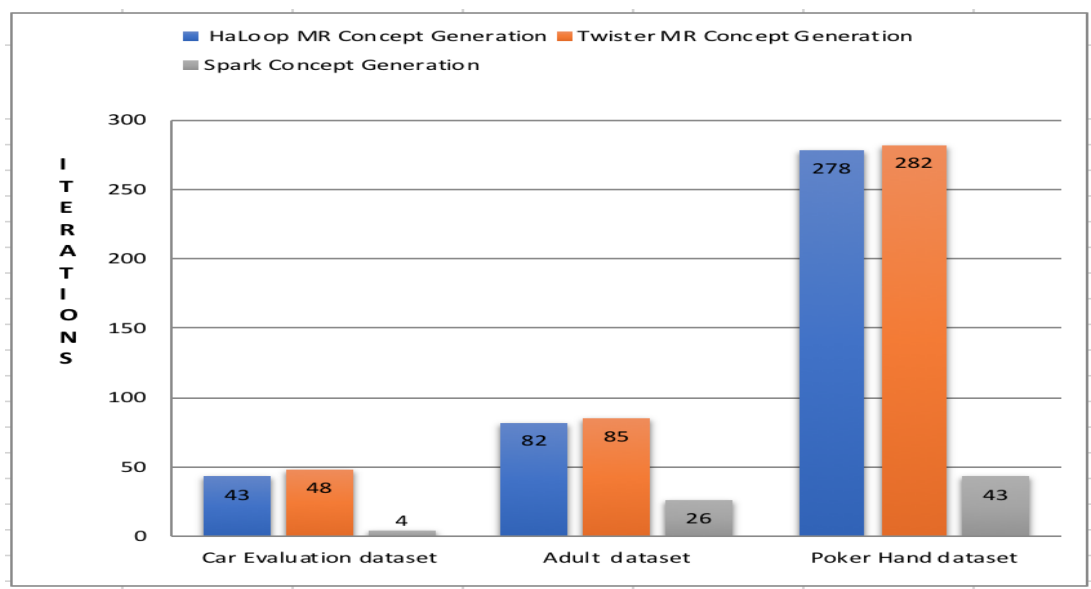

**Fig. 15** Number of iterations that Spark, HaLoop and Twister based environments taken for execution of each dataset on 16 node cluster

**Table 11** Number of iterations taken by Spark and Hadoop approaches

| Algorithm | Car Evaluation dataset | Adult dataset | Poker Hand dataset |
|---|---|---|---|
| HaLoop MR Concept Generation | 43 | 82 | 278 |
| Twister MR Concept Generation | 48 | 85 | 282 |
| Spark Concept Generation | 4 | 26 | 43 |

The below graph in Fig 17 shows the running time of the algorithm for each iteration. During the first iterations, both Spark and MapReduce approaches took almost same time. Spark completed later iterations in very short time because of storing all the data in the cache memory and reusing the cached data. The below graph in Fig 14 shows the time taken for first iterations on Poker-Hand dataset. In the MapReduce approaches, every iteration involves disk access for reading and writing output to disk after map and reduce phases, shuffling and sorting operations. Because of this reason, the time it takes to complete a particular iteration is high which is eventually increasing the running time of the application. In Spark only the first iteration is taking time because of reading input data from disk. From later iterations Spark always tries to take data from cache. This is helpful for low processing times of iteration while executing data on large datasets.

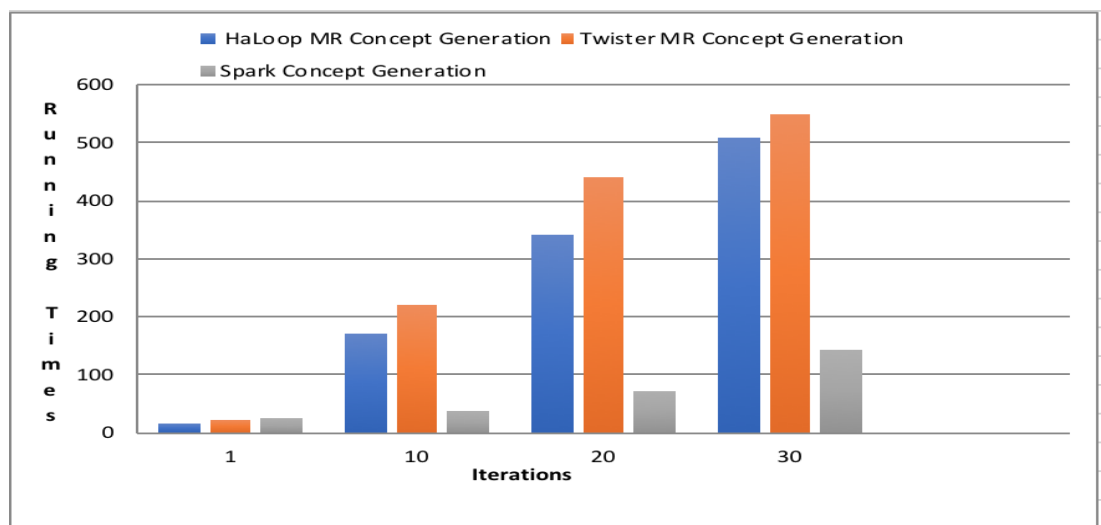

**Fig. 16** Running time of algorithms under different iteration values

We also tried crashing a node while the job was running on car evaluation dataset after ten iterations on a four node cluster; this slows down the job by 44 seconds (20% on average). The data partitions on the lost node are recomputed and cached parallel to other nodes quickly with the help of lineage graph. The part of the lineage graph of the SparkConcept algorithm is given below in Fig.
18. During the stage 0 the file was read from disk and converted to an RDD. Various transformations and actions are performed on the transformed RDD as part of concept generation process can be seen in stage 4 and stage 5.

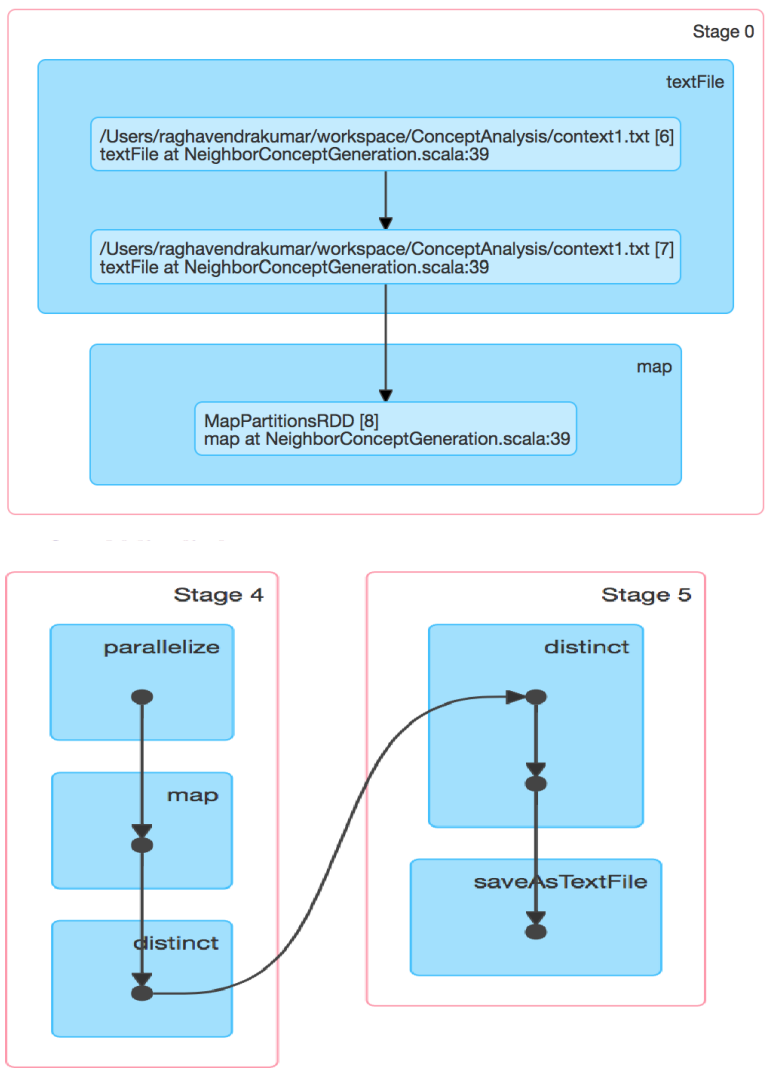

**Fig. 17** Part of the lineage graph from the SparkConceptGeneration algorithm

# 7 Conclusion

There are several works proposed for the generation of concepts and lattice graph construction. The work proposed in this paper can be seen as evidence of computing formal concepts and constructing lattice graph by isolated nodes. The main drawback of the existing distributed algorithms is acquiring hard- ware with several processor cores, efficiently using system resources like memory, CPU etc.. and handling fault tolerance effectively. Also the existing batch processing applications do not provide the lattice structure after the concept generation. The proposed model overcomes all the drawbacks and effectively utilized system resources and builds the digraph of the concepts efficiently. The in-memory computations in apache Spark helped in generating concepts more quickly and the two step canonicity eliminated the processing of duplicate concepts. With all these benefits, the experimental analysis conducted on the proposed model also proves that it is working better than the other existing distributed approaches. Furthermore, the core idea behind the implementation of these algorithms is to employ FCA in various domains to get effective results. Our future work will focus on concept reduction, construction of concept neighborhoods, and using the proposed algorithms in various applications that requires large input.